# ИНДЕКС ТОНАЛЬНОСТИ РУССКОЯЗЫЧНОГО ФЕЙСБУКА

**Панченко А. И.** (alexander.panchenko@uclouvain.be)

Лувенский католический университет, Лувен, Бельгия;
ООО «Лаборатория Цифрового Общества», Москва, Россия

Индекс тональности измеряет эмоциональный уровень в корпусе текстов. В данной работе мы определяем четыре подобных индекса. Предложенные индексы используются для измерения общего уровня «позитивности» группы пользователей на основании их постов в социальной сети. В отличие от предыдущих работ, мы впервые вводим индексы, которые работают на уровне текстов, а не отдельных слов. Кроме этого, данная работа впервые представляет результаты вычисления индекса тональности на значительной части русскоязычного Фейсбука. Полученные результаты согласуются с результатами подобных экспериментов на англоязычных корпусах.

**Ключевые слова:** анализ тональности текста, индекс тональности, Фейсбук

# SENTIMENT INDEX OF THE RUSSIAN SPEAKING FACEBOOK

**Panchenko A. I.** (alexander.panchenko@uclouvain.be)

Universite catholique de Louvain, Louvain-la-Neuve, Belgium;
Digital Society Laboratory LLC, Moscow, Russia

A sentiment index measures the average emotional level in a corpus. We introduce four such indexes and use them to gauge average "positiveness" of a population during some period based on posts in a social network. This article for the first time presents a text-, rather than word-based sentiment index. Furthermore, this study presents the first large-scale study of the sentiment index of the Russian-speaking Facebook. Our results are consistent with the prior experiments for English language.

## 1. Introduction

Social media analysis has opened new exciting possibilities for social and economic sciences [14]. In this paper, we present a technique for measuring *social sentiment index*. This metric reflects the emotional level of a social group by means of text analysis. As Dodds et al. [5] notice, "a measure of societal happiness is a crucial adjunct to traditional economic measures such as gross domestic product".

Social sentiment index is based on the sentiment analysis technology. There is a huge number of papers on sentiment analysis, most notable of English [1,6,7,8], but also of Russian [9, 10] texts. However, most of these studies focus on classification of a single text that expresses sentiment about a particular entity. Some attempts were also made to calculate sentiment indexes. For instance, Godbole et al. [1] performed sentiment analysis of news and blogs and computed a sentiment index of named entities, such as George Clooney or Slobodan Milosevic.

Several researchers tried to measure "positiveness" or "happiness" of social network users. Most experiments with the happiness indexes were conducted on English texts. Several works dealt with the emotionally-annotated posts from LiveJournal, the biggest online community back to 2005. Mihalchea and Lui [12] analyzed some 10,000 posts from LiveJournal, annotated with the tags "happy" or "sad". The authors identified terms that characterize happiest and saddest texts. According to this study, the happiest day is Saturday, while the happiest hours are 3 am and 9 pm. Mishne and Rijke [17] performed another experiment on the LiveJournal data. The authors collected a corpus of 8 million posts labeled with 132 distinct mood tags such as "sad", "drunk" or "happy". This dataset was used to train supervised models predicting tags by texts. In particular, this work studied feedback of the blogosphere on a terror attack in London. Balog and Rijke [18] performed a time series analysis of 20 million LiveJournal posts issued during one year (2005–2006). The researchers observed a clear impact of the weather, holidays and season on the mood of some profiles.

More recent studies are focused on analysis of social networks, such as Twitter and Facebook. Dodds et al. [5] analyzed 4.6 billion of tweets over a period from 2008 to 2011. According to the authors, the dataset represented 5% of Twitter back to 2011. The researchers measured "happiness" with a simple dictionary-based approach and came to several conclusions: the happiest day of the week is Saturday; the happiest hour is 5 am; peaks of happiness coincide with national holidays, such as Christmas and St. Valentine's day.

Kamvar and Harris [15] developed a system that maintains an index of 14 million phrases of 2.5 million people starting from "I feel" or "I am feeling". The phrases along with information about their respective authors were crawled from blogs, microblogs and social networks. The system is able to answer questions like "Do Europeans feel sad more often than Americans?" or "How did young people in Ohio feel when Obama was elected?".

The work of Kramer [13] is probably the most similar to our experiment. The author analyzes use of emotion words by roughly 100 million Facebook users since 2007. He proposed a "Gross National Happiness" index, based on a standardized difference between positive and negative terms in texts. The peaks of this index coincide with the nationally important dates.

Recent work of Wang et al. [16] challenges the idea of the Gross National Happiness index by comparing its results to the data gathered with a Facebook user questionnaire "myPersonality"[1]. The authors did not find a significant correlation between self-reported happiness and happiness computed from their posts.

Two contributions of this paper to the exploration of sentiment/happiness indexes are as follows:
1. We propose four indexes that gauge emotional level of population. Unlike most previous works, we not only deal with word-based indexes, but also propose a text-based sentiment index.
2. To the best of our knowledge, this is the first study of sentiment index of the Russian-speaking social network.

## 2. Social Sentiment Index

First, this section presents several network datasets used to calculate social sentiment indexes. Next, we describe sentiment analysis method that is used in calculations. Finally, we present four indexes used in our experiment.

### 2.1. The Facebook Corpus

We use in our experiments a corpus of Facebook posts provided for research purposes by Digsolab LLC[2]. This anonymized dataset contains texts from the publically available part of the social network. According to our collaborator at Digsolab, the corpus was collected using the API of Facebook[3]. Construction of such anonymized samples from Facebook is common in the research community [12]. The dataset we deal with contains 573 million anonymized posts and comments of 3,190,813 users. We considered only texts in Russian. The language detection was completed using the *langid.py* module [2]. Table 1 summarizes key parameters of the dataset.

The oldest post in the corpus dates back to the 5[th] August of 2006, while the latest is of 13[th] November 2013. The distribution of posts over time is far from being uniform (see Fig. 1). As one may expect, the number of posts in the social network grows rapidly as the network gets popular with Russian-speaking users. The biggest number of texts per day (1,243,310) was observed on the 4-th April 2013. After this peak, the number of posts drops to 600–700 thousand per day. This is an artifact of the data collection method used. However, as sentiment index is a relative value (see Section 2.3) it should not be much affected by such noise.

---

[1]  http://www.psychometrics.cam.ac.uk/productsservices/mypersonality

[2]  http://www.digsolab.com/

[3]  https://developers.facebook.com/tools/explorer?method=GET&path=EuropeanCommission%2Fposts

**Table 1.** Statistics of the Facebook corpus

| Number of anonymized users | 3,190,813 |
|---|---|
| Language | Russian |
| Number of posts | 426,089,762 |
| Number of comments | 147,140,265 |
| **Number of texts (posts + comments)** | **573,230,027** |
| Number of tokens in posts | 20,775,837,467 |
| Number of tokens in comments | 2,759,777,659 |
| **Number of tokens (posts + comments)** | **23,535,615,126** |
| Average post length, tokens | 49 |
| Average comment length, tokens | 19 |

Facebook claims [3] that by March 2013 it had about 1.19 billion active users. According to Internet World Stat [4], the number of Facebook users in Russia by the end of 2012 is 7,963,400. Thus, our sample is roughly equal to 40% of the 2012 Russian Facebook.

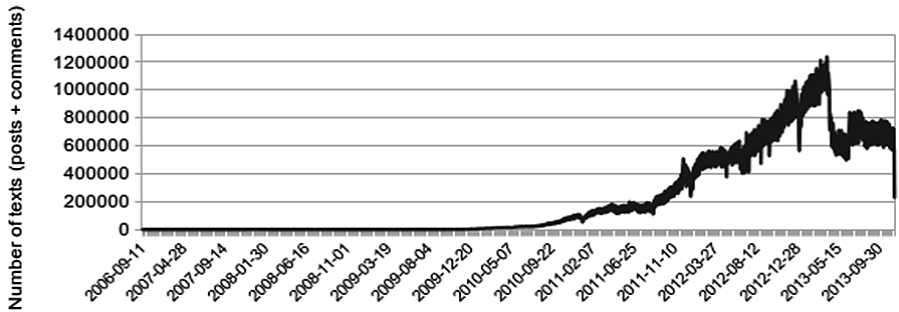

**Fig. 1.** Number of texts per day in the Facebook corpus

### 2.2. Sentiment Detection Method

The social sentiment index builds upon a popular approach to sentiment analysis, based on a dictionary of positive and negative terms [6,7]. Similar approach was employed by Dodds et al. [5] to measure "happiness" of Twitter users.

We use a custom sentiment dictionary developed specifically for the social media texts. Two annotators independently labeled 15,000 most frequent adjectives in the aforementioned corpus. The result is a dictionary of 1,511 terms, where each term is supposed to be a context-independent predictor of a positive/negative opinion. Each term was labeled by both annotators as positive or negative. This dictionary contains 600 negative terms ($D_-$) and 911 negative terms ($D_+$) such that each term $w \in \{D_- \cup D_+\}$ was labeled as positive or negative by both annotators. Table 2 presents some examples of these terms.

**Table 2.** Most frequent positive and negative adjectives in the Facebook corpus

| Positive adjectives | | Negative adjectives | |
|---|---|---|---|
| хороший | good | плохой | bad |
| новый | new | старый | old |
| первый | first | долгий | long |
| нужный | helpful | неблагоприятный | unfavorable |
| бесплатный | free of charge | скучный | boring |
| любимый | beloved | сложный | complicated |
| интересный | interesting | голодный | hungry |
| спокойный | quiet | страшный | scary |
| социальный | social | скучно | bored |
| добрый | kind | немой | mute |

Each text $t$ is represented as a multiset of $n$ lemmas $\{w_1, \ldots, w_n\}$. The lemmatization was done with the morphological analyzer *PyMorphy*[4]. We used the following decision rule to classify a text $t$ as positive (+1), negative (−1) or neutral (0):

$$c(t) = \begin{cases} +1 & if \quad |\Delta| \geq \alpha \ and \ \Delta > 0 \\ 0 & if \quad |\Delta| < \alpha \\ -1 & if \quad |\Delta| \geq \alpha \ and \ \Delta < 0 \end{cases}, \quad where \ \Delta = \frac{|\{w \in t \mid w \in D_+\}| - |\{w \in t \mid w \in D_-\}|}{n}$$

Here the *emotion delta* $\Delta \in [-1; 1]$ shows both direction and amplitude of a text sentiment. According to the decision rule mentioned above, a text is classified as neutral if $\Delta$ is less than some constant $\alpha$. In our experiments, we fixed $\alpha = 0.05$. This value provides a good tradeoff between precision and recall (see Table 3).

We deliberately avoided machine learning, as Facebook is an open domain corpus. Statistical sentiment classifiers trained on one domain are known to perform poorly on another domain [6,7], as such classifiers can heavily rely on terms that are positive or negative only within a specific field. According to Pang and Lee [6], "simply applying the classifier learned on data from one domain barely outperforms the baseline for another domain". Therefore, we rather opted for a dictionary-based approach, that relies on a set of *frequent domain-invariant* positive and negative terms.

We compared the dictionary-based classification method with the baselines on the ROMIP 2012 dataset [10]. This dataset contains some 50 thousands of reviews about films, movies and digital cameras. Results of this evaluation are presented in Table 3.

---

[4]  https://bitbucket.org/kmike/pymorphy

**Table 3.** Performance of the dictionary-based sentiment classification approach, as compared to other methods (ROMIP-2012 dataset)

| RunID | Object | Macro_P | Macro_R | Macro_F1 | Accuracy | P_1 | P_0 | P_-1 | R_1 | R_0 | R_-1 |
|---|---|---|---|---|---|---|---|---|---|---|---|
| xxx | books | 0.379 | 0.443 | 0.350 | 0.536 | 0.873 | 0.157 | 0.106 | 0.604 | 0.157 | 0.569 |
| sentistrength | books | 0.368 | 0.403 | 0.327 | 0.448 | 0.848 | 0.154 | 0.103 | 0.468 | 0.375 | 0.367 |
| yyy | books | 0.399 | 0.494 | 0.377 | 0.560 | 0.908 | 0.154 | 0.136 | 0.620 | 0.183 | 0.678 |
| nb-blinov | books | 0.408 | **0.528** | **0.390** | **0.675** | 0.909 | 0.157 | 0.157 | 0.785 | 0.042 | **0.757** |
| dict (α = 0.02) | books | 0.42 | 0.431 | 0.348 | 0.445 | 0.893 | **0.175** | 0.191 | 0.42 | 0.677 | 0.197 |
| dict (α = 0.05) | books | 0.437 | 0.404 | 0.274 | 0.327 | 0.919 | 0.163 | 0.229 | 0.246 | 0.844 | 0.122 |
| dict (α = 0.07) | books | **0.446** | 0.381 | 0.217 | 0.261 | **0.934** | 0.157 | 0.248 | 0.155 | **0.911** | 0.077 |
| Xxx | movies | 0.395 | 0.454 | 0.361 | 0.493 | 0.819 | 0.235 | 0.131 | 0.586 | 0.148 | 0.628 |
| Sentistrength | movies | 0.371 | 0.401 | 0.343 | 0.436 | 0.774 | 0.219 | 0.119 | 0.485 | 0.274 | 0.445 |
| Yyy | movies | 0.411 | **0.497** | 0.390 | 0.522 | 0.849 | 0.221 | 0.165 | 0.610 | 0.173 | **0.708** |
| nb-blinov | movies | 0.347 | 0.489 | **0.400** | **0.705** | 0.788 | 0.000 | 0.253 | **0.943** | 0.000 | 0.525 |
| dict (α = 0.02) | movies | 0.453 | 0.438 | 0.383 | 0.465 | 0.852 | 0.264 | 0.241 | 0.416 | 0.723 | 0.177 |
| dict (α = 0.05) | movies | 0.473 | 0.412 | 0.315 | 0.382 | 0.892 | **0.249** | 0.278 | 0.259 | 0.869 | 0.109 |
| dict (α = 0.07) | movies | **0.48** | 0.388 | 0.26 | 0.329 | **0.908** | 0.239 | **0.293** | 0.172 | **0.923** | 0.069 |
| Xxx | cameras | 0.388 | 0.390 | 0.370 | 0.561 | 0.864 | 0.111 | 0.190 | 0.632 | 0.138 | 0.400 |
| Sentistrength | cameras | 0.373 | 0.359 | 0.319 | 0.429 | 0.855 | 0.106 | 0.157 | 0.461 | 0.370 | 0.247 |
| Yyy | cameras | 0.445 | **0.488** | **0.443** | 0.628 | 0.903 | **0.127** | 0.303 | 0.687 | 0.312 | **0.464** |
| nb-blinov | cameras | 0.445 | 0.441 | 0.437 | **0.816** | 0.844 | 0.000 | **0.490** | 0.974 | 0.000 | 0.350 |
| dict (α = 0.02) | cameras | 0.452 | 0.359 | 0.294 | 0.411 | 0.889 | 0.102 | 0.365 | 0.439 | 0.564 | 0.073 |
| dict (α = 0.05) | cameras | 0.441 | 0.329 | 0.193 | 0.256 | 0.905 | 0.095 | 0.324 | 0.232 | 0.733 | 0.022 |
| dict (α = 0.07) | cameras | **0.462** | 0.32 | 0.141 | 0.181 | **0.912** | 0.093 | 0.381 | 0.13 | **0.816** | 0.015 |

Our method is denoted as *dict*. The table also features performances of four alternative sentiment classification methods. The *xxx* and *yyy* are commercial sentiment classification systems[5]. The system *sentistrength* is a freely available tool for research purposes, developed by Thelwall et al. [8]. In this benchmark, we used standard Russian dictionaries coming with this tool with default parameters. The system *nb-blinov* is based on the sentiment phrase dictionary developed by Blinov et al. [9]. The dictionary specifies conditional probabilities of 19,000 phrases given a positive or negative class, e.g.:

$$p(w|-1) = p(\text{еле досматривать}|-1) = 0.000881168;$$
$$p(w|+1) = p(\text{еле досматривать}|+1) = 0.000016001.$$

The method *nb-blinov* relies on the decision rule, used in the Naïve Bayes classifier:

$$c(t) = \arg\max_{c \in \{-1, 0, +1\}} p(c) \prod_{\{w \in T \mid w: D_+ \cup D_-\}} p(w|c)$$

Table 3 presents results of the comparison in terms of precision (Macro_P, P_-1, P_0, P_1), recall (Macro_R, R_-1, R_0, R_1), F-measure (Macro_F1) and accuracy.

---

[5] We cannot reveal their names, as the developers who provided the demo-versions did not let us do so. The *xxx* relies on both machine learning and rules, while *yyy* is a rule-based system.

As we may observe, the precision of the dictionary-based technique on the positive and negative class matches or even outperforms precision of the other systems. On the other hand, the recall of the *dict* method is significantly lower as compared to other techniques. As one may expect, the bigger the value of the parameter the higher the precision and the lower the recall.

We conclude that the dictionary-based method described in Section 2.2 is suitable for the needs of the social sentiment index, as it is domain-independent and yields the baseline performance.

### 2.3. Sentiment Indexes

The goal of a sentiment index is to measure positiveness of social network users during some period. For instance, one may want to know if high values of the index are related to national holidays, such as New Year, while low values of the index occur during "depressing" periods related to national tragedies, such as air crashes. Sentiment index can be useful (1) to analyze feedback of social media users on notable events and (2) for prediction of "planned" peaks of positiveness/negativeness. In our experiment, we used the following four simple metrics:

1. **Word Sentiment Index** is a ratio of positive to negative terms in all texts (posts and comments) in a corpus *T*:

$$s_w = \frac{\sum_{t \in T} |\{w \in t \mid w \in D_+\}| + \epsilon}{\sum_{t \in T} |\{w \in t \mid w \in D_-\}| + \epsilon}$$

2. **Text Sentiment Index** is a ratio of positive to negative texts in the corpus *T*:

$$s_t = \frac{|\{t \in T \mid c(t) = +1\}| + \epsilon}{|\{t \in T \mid c(t) = -1\}| + \epsilon}$$

3. **Word Emotion Index** is a ratio of emotional (positive or negative) words in texts in the corpus *T*:

$$e_w = \frac{\sum_{t \in T} |\{w \in t \mid w \in \{D_+ \cup D_-\}\}| + \epsilon}{\sum_{t \in T} |t| + \epsilon}$$

4. **Text Emotion Index** is a ratio of emotional (positive or negative) texts in the corpus *T*:

$$e_t = \frac{|\{t \in T \mid c(t) = \{-1, +1\}\}| + \epsilon}{|T| + \epsilon}$$

Both text and word sentiment indexes are in the range $[0; +\infty]$; here $\epsilon$ was set to $10^{-6}$. The first index works on the lexical level, while the second index deals with texts. Thus, precision of the second index depends on precision of the decision rule *c(t)*. Finally, the last two indexes capture the overall emotional level in a social network.

## 3. Results and Discussion

Table 4 presents results of the sentiment analysis of the entire corpus. According to both word-, and text-based indexes, positiveness predominate over negativeness. Users employ positive terms approximately 3.8 times more often than the negative terms. People use more emotional terms in the comments and fewer positive or negative terms in posts. Our results show that, during the entire period, people issued roughly 7.5 times more positively oriented texts than negatively oriented texts. Thus, we see that users of the Facebook tend to write much more positive texts, than negative ones. This trend is common for both posts and comments.

Indeed, people are keen to share happy moments and are reluctant to share dramas in their lives. Furthermore, by analyzing only public data, we probably miss a significant share of posts with a negative sentiment. One can expect that users tend to discuss negative matters in close circle of friends.

**Table 4.** Results of the social sentiment index calculation (the entire period)

|  | posts, % | comments, % | posts + comments, % |
|---|---|---|---|
| positive words | 1.38 | 2.06 | 1.72 |
| negative words | 0.37 | 0.47 | 0.42 |
| Word Emotion Index, $e_w$ | 1.75 (0.017) | 2.53 (0.025) | 2.14 (0.021) |
| Word Sentiment Index, $s_w$ | 3.72 (0.037) | 4.38 (0.044) | 3.81 (0.038) |
| positive texts | 13.43 | 18.42 | 14.71 |
| negative texts | 1.83 | 2.28 | 1.94 |
| Text Emotion Index, $e_t$ | 15.26 (0.153) | 20.70 | 16.65 (0.166) |
| Text Sentiment Index, $s_t$ | 7.34 (0.073) | 8.08 | 7.58 (0.076) |

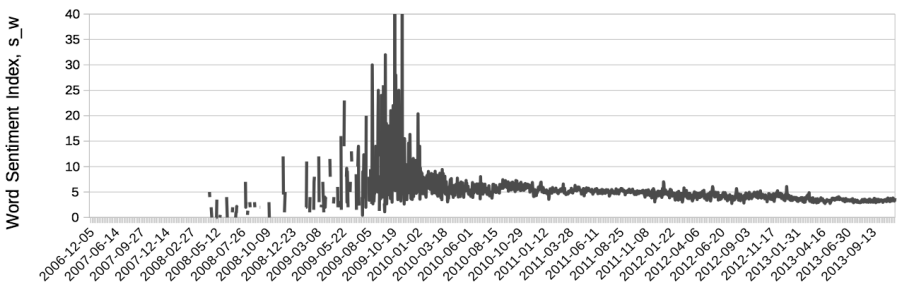

**Fig. 2.** Evolution of the word sentiment index $s_w$ during the entire period

Fig. 2 shows dynamics of the word sentiment index $s_w$ during the entire period. First, values of the index are significantly greater than one most of the time. Second, the index is fluctuating greatly from day to day. As we already learned from Fig. 1, most texts in the dataset were issued after the 1-st January of 2011. We suppose that, the huge fluctuations of the sentiment index from 2006 until the end of 2010 are due to sparseness/incompleteness of the data. Therefore, we will focus the further analysis on the last two years: 01/11/2011–31/10/2012 and 01/11/2012–31/10/2013.

Fig. 3 presents the sentiment indexes for the two last years. The index fluctuates a lot with time: dynamic range of the word sentiment index is 5.5 (min/max), while dynamic range of the text sentiment index is 6. We used day-based plots as we found that week- and month-based plots are less informative. As one may notice, positive/negative bursts are short and thus smoothed easily. The extreme values of the indexes indeed coincide with the some important events in Russia and in the World:

(1) 31-12-*—New Year (+);
(2) 14-02-*—St.Valentine's day (+);
(3) 23-02-*—Man's day (+);
(4) 08-03-*—Woman's day (+);
(5) 09-05-*—Victory Day, World War 2 commemorative day (–);
(6) 07-07-2012—Krasnodar Krai floods in Russia[6] (–);
(7) 22-07-2012—A new unpopular law regulating non-profit organizations in Russia[7] (–);
(8) 16-09-2012—A mass protest against government in Russia[8] (–);
(9) 25-10-2012—Hurricane Sandy in US (–).

The list above merely reflects our interpretation of the sentiment index. While points (1)–(5) seem to be correct, extreme values of the index around points (6)–(9) can be due to a different cause. We did not have resources to validate our interpretations, as it would require manual check of a huge number of texts. For instance, to verify point (7) one would need to check topic of 17,490 texts issued on the 22/07/2012. However, it is clear that further interpretation of the results is desirable.

Fig. 4 depicts the emotional indexes for the two last years. The fraction of emotional texts $e_t$ ranges between 3% and 5%. This quantity is relatively stable and has practically no huge bursts. The fraction of emotional words $e_w$ varies in a similar way: it ranges from 0.15% until 0.35%. Dynamic range of both emotional indexes is about 2.2. One can observe a significant discrepancy of $e_t$ and $e_w$ during the last two years (Fig. 3) and during the whole period (Fig. 2). These differences are again due to the noisiness of the data during the period 2006–2010 (see Fig. 1 and 2). Some further observations are as follows:

- Emotional level of the social network varies periodically. The word and text emotional indexes ($e_w$ and $e_t$) vary weekly (see Fig. 4 and 5). Fewer text are issued during weekends; fewer *emotional* texts are issued during weekends.
- Direction of emotions varies irregularly (see Fig. 3); extreme values of word and text sentiment indexes ($s_w$ and $s_t$) coincide with some important events, such as New Year or a huge natural disaster.

---

[6] http://en.wikipedia.org/wiki/2012_Krasnodar_Krai_floods

[7] http://rusnewsjournal.com/menu/186/

[8] http://www.nytimes.com/2012/09/16/world/europe/anti-putin-protesters-march-in-moscow-russia.html?_r=0

It is clear that grouping results by sociodemographic factors can be of interest. This will let approach questions such as "If people from Moscow are in average more positive than people from Saint Petersburg?" or "If older people are more positive than youth?", etc. However, in our pilot study we limit ourselves to measuring "global" indexes that average sentiment of all users.

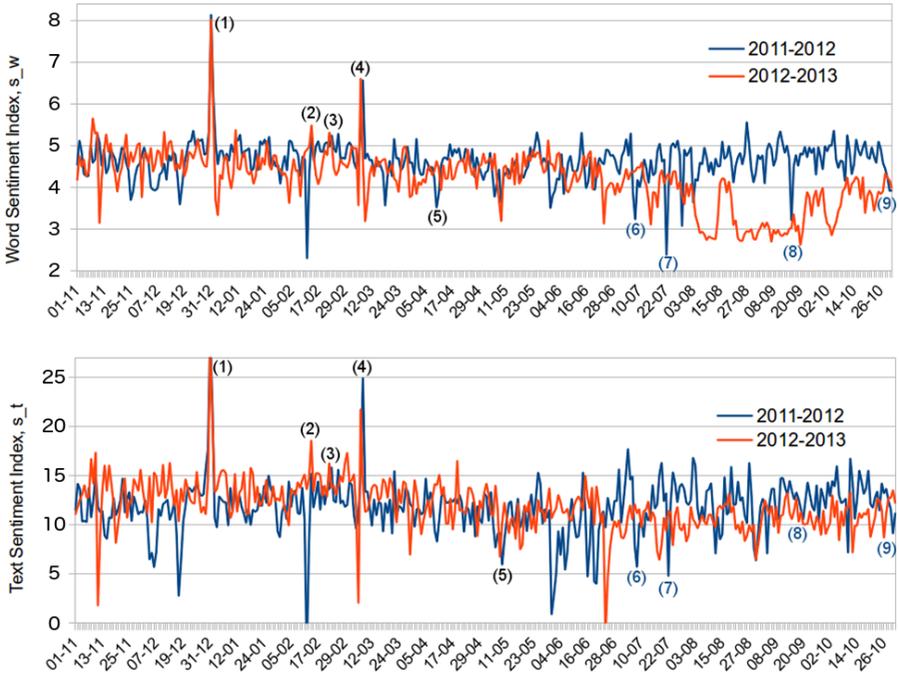

**Fig. 3.** Word ($s_w$) and text ($s_t$) sentiment indexes temporal evolution during two years: 11/2011—11/2012 and 11/2012—11/2013

## 4. Conclusion and Future Work

This paper presented some preliminary results on the social sentiment index of the Russian-speaking Facebook. First, we introduced four indexes that measure average emotional level in a social network. Two of them rely on sentiment words, while two other exploit text sentiment classification. We applied the proposed metrics on a corpus of Facebook posts. According to our results: (1) positive posts and comments predominate over negative ones; (2) maximum values of the index coincide with the national holidays, while minimum values coincide with national tragedies and commemoration days.

In further work, we plan to introduce a more sophisticated version of the index taking into account other factors, such as influence of a post and its author. We also would like to perform further analysis of the results to answer the questions such as: "Which time of the day is the most positive?" or "Are women use more positive terms than men?".

Panchenko A. I.

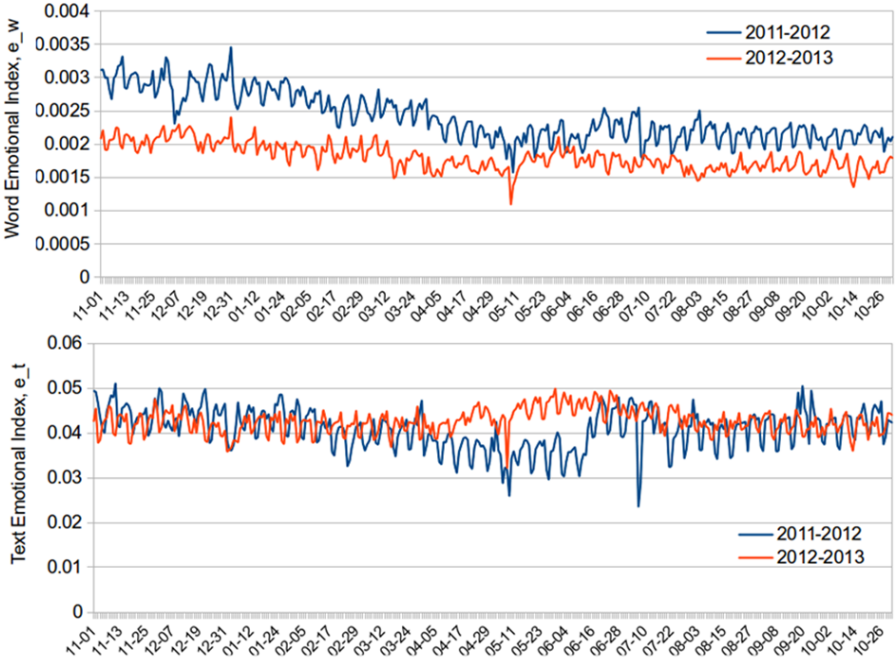

**Fig. 4.** Word ($e_w$) and text ($e_t$) emotional indexes temporal evolution during the two years: 11/2011—11/2012 and 11/2012—11/2013

## Acknowledgements

This research was supported by Digital Society Laboratory LLC. We thank Sergei Objedkov and three anonymous reviewers for their helpful comments that significantly improved quality of this paper.

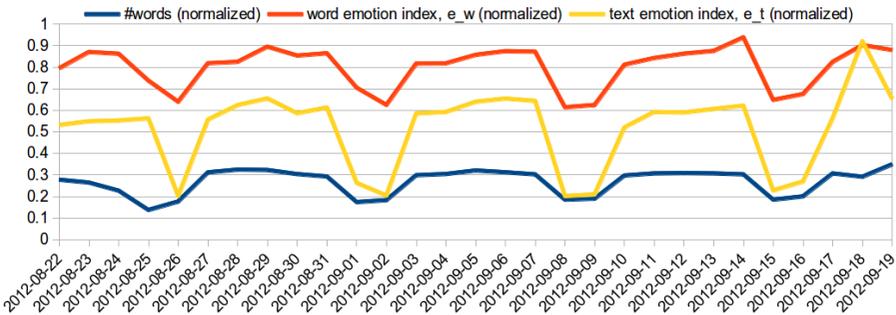

**Fig. 5.** Word ($e_w$) and text ($e_t$) emotional indexes, compared to the number of words. Each variable was projected to the range [0;1]